\documentclass[10pt,twocolumn,letterpaper]{article}

\usepackage{cvpr}              

\usepackage{graphicx}
\usepackage{amsmath}
\usepackage{amssymb}
\usepackage{booktabs}

\usepackage{color}
\usepackage{xcolor}

\definecolor{orange_bar}{RGB}{244, 177, 131}
\definecolor{grey_bar}{RGB}{132, 151, 176}

\usepackage[pagebackref,breaklinks,colorlinks]{hyperref}

\usepackage[capitalize]{cleveref}
\crefname{section}{Sec.}{Secs.}
\Crefname{section}{Section}{Sections}
\Crefname{table}{Table}{Tables}
\crefname{table}{Tab.}{Tabs.}

\newcommand{\para}[1]{\vspace{1.2ex}\noindent\textbf{#1}\hspace{1ex}}

\def\cvprPaperID{3016} 
\def\confName{CVPR}
\def\confYear{2023}

\begin{document}

\title{MOTRv2: Bootstrapping End-to-End Multi-Object Tracking by Pretrained Object Detectors}

\author{
  Yuang Zhang\textsuperscript{\rm 1*},
  Tiancai Wang\textsuperscript{\rm 2},
  Xiangyu Zhang\textsuperscript{\rm 2,3}\\
  \textsuperscript{\rm 1}Shanghai Jiao Tong University\hspace{0.6em}
  \textsuperscript{\rm 2}MEGVII Technology\hspace{0.6em}
  \textsuperscript{\rm 3}Beijing Academy of Artificial Intelligence 
}

\maketitle

{\let\thefootnote\relax\footnotetext{* The work was done during internship at MEGVII Technology and supported by National Key R\&D Program of China (2020AAA0105200) and Beijing Academy of Artificial Intelligence (BAAI).}}

\begin{abstract}
In this paper, we propose MOTRv2, a simple yet effective pipeline to bootstrap end-to-end multi-object tracking with a pretrained object detector. Existing end-to-end methods, \eg MOTR~\cite{zeng2022motr} and TrackFormer~\cite{Meinhardt2021trackformer} are inferior to their tracking-by-detection counterparts mainly due to their poor detection performance.  We aim to improve MOTR by elegantly incorporating an extra object detector.
We first adopt the anchor formulation of queries and then use an extra object detector to generate proposals as anchors, providing detection prior to MOTR.
The simple modification greatly eases the conflict between joint learning detection and association tasks in MOTR. MOTRv2 keeps the query propogation feature and scales well on large-scale benchmarks. 
MOTRv2 ranks the 1st place (\textbf{73.4\%} HOTA on DanceTrack) in the 1st Multiple People Tracking in Group Dance Challenge.
Moreover, MOTRv2 reaches state-of-the-art performance on the BDD100K dataset. We hope this simple and effective pipeline can provide some new insights to the end-to-end MOT community. Code is available at \url{https://github.com/megvii-research/MOTRv2}.
\end{abstract}

\section{Introduction}
\label{sec:intro}
Multi-object tracking (MOT) aims to predict the trajectories of all objects in the streaming video. It can be divided into two parts: detection and association. For a long time, the state-of-the-art performance on MOT has been dominated by tracking-by-detection methods~\cite{bewley2016simple,zhang2020fairmot, zhang2021bytetrack, wang2019jde} with good detection performance to cope with various appearance distributions. These trackers~\cite{zhang2021bytetrack} first employ an object detector (e.g., YOLOX~\cite{yolox2021}) to localize the objects in each frame and associate the tracks by ReID features or IoU matching.
The superior performance of those methods partially results from the dataset and metrics biased towards detection performance. However, as revealed by the DanceTrack dataset~\cite{peize2021dance}, their association strategy remains to be improved in complex motion.

\begin{figure}[t]
  \centering
  \includegraphics[width=\linewidth]{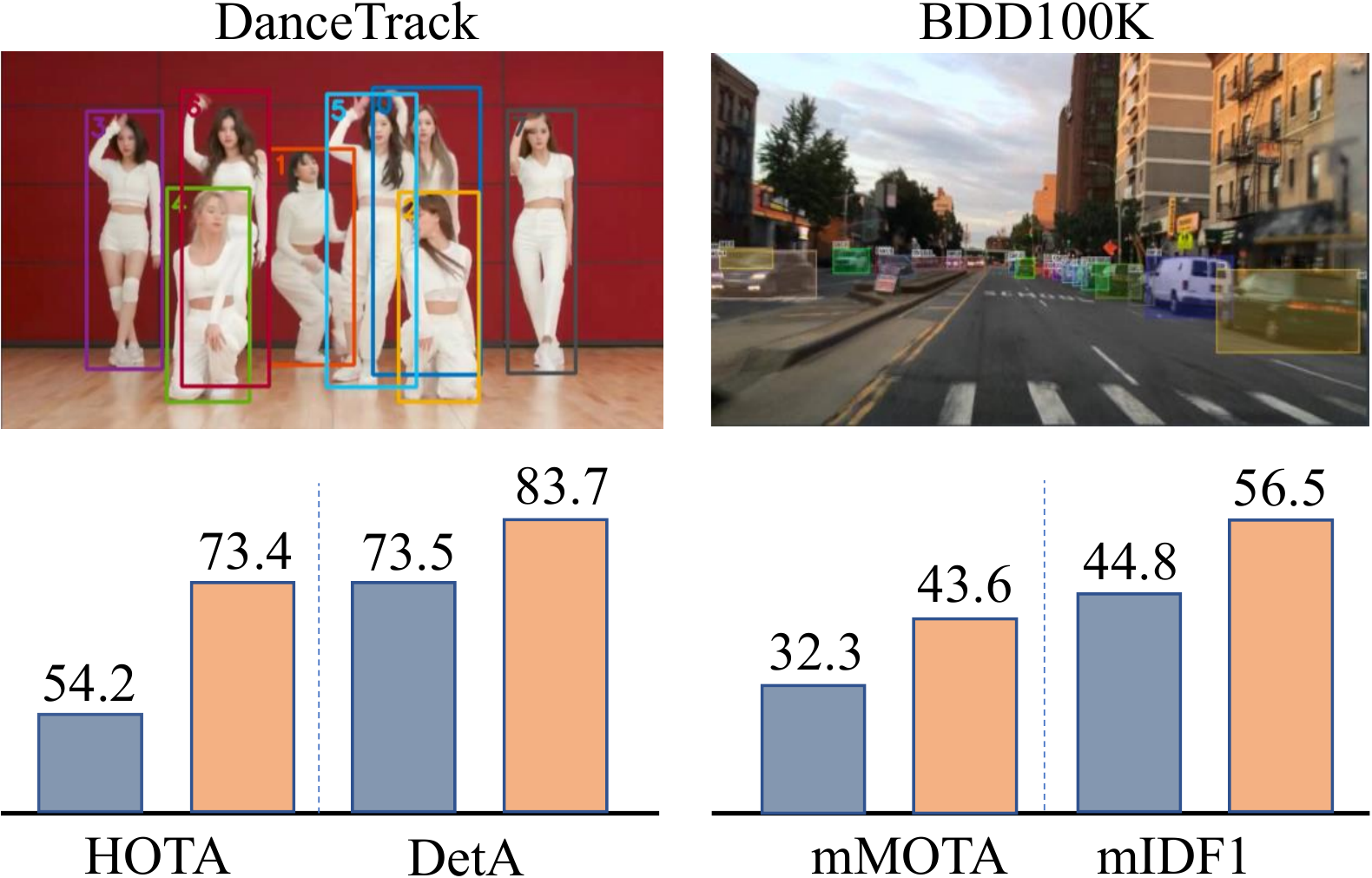}
  \caption{Performance comparison between MOTR (\textcolor{grey_bar}{grey bar}) and MOTRv2 (\textcolor{orange_bar}{orange bar}) on the DanceTrack and BDD100K datasets. MOTRv2 improves the performance of MOTR by a large margin under different scenarios.}
  \vspace{-2ex}
  \label{fig:comparewithv1}
\end{figure}

Recently, MOTR~\cite{zeng2022motr}, a fully end-to-end framework is introduced for MOT. The association process is performed by updating the tracking queries while the new-born objects are detected by the detect queries. Its association performance on DanceTrack is impressive while the detection results are inferior to those tracking-by-detection methods, especially on the MOT17 dataset. We attribute the inferior detection performance to the conflict between the joint detection and association processes. Since state-of-the-art trackers \cite{chu2021transmot,zhang2021bytetrack,cao2022observation} tend to employ extra object detectors, one natural question is how to incorporate MOTR with an extra object detector for better detection performance. One direct way is to perform IoU matching between the predictions of track queries and extra object detector (similar to TransTrack~\cite{transtrack}). In our practice, it only brings marginal improvements in object detection while disobeying the end-to-end feature of MOTR.

\begin{figure*}
  \centering
  \includegraphics[width=0.95\linewidth]{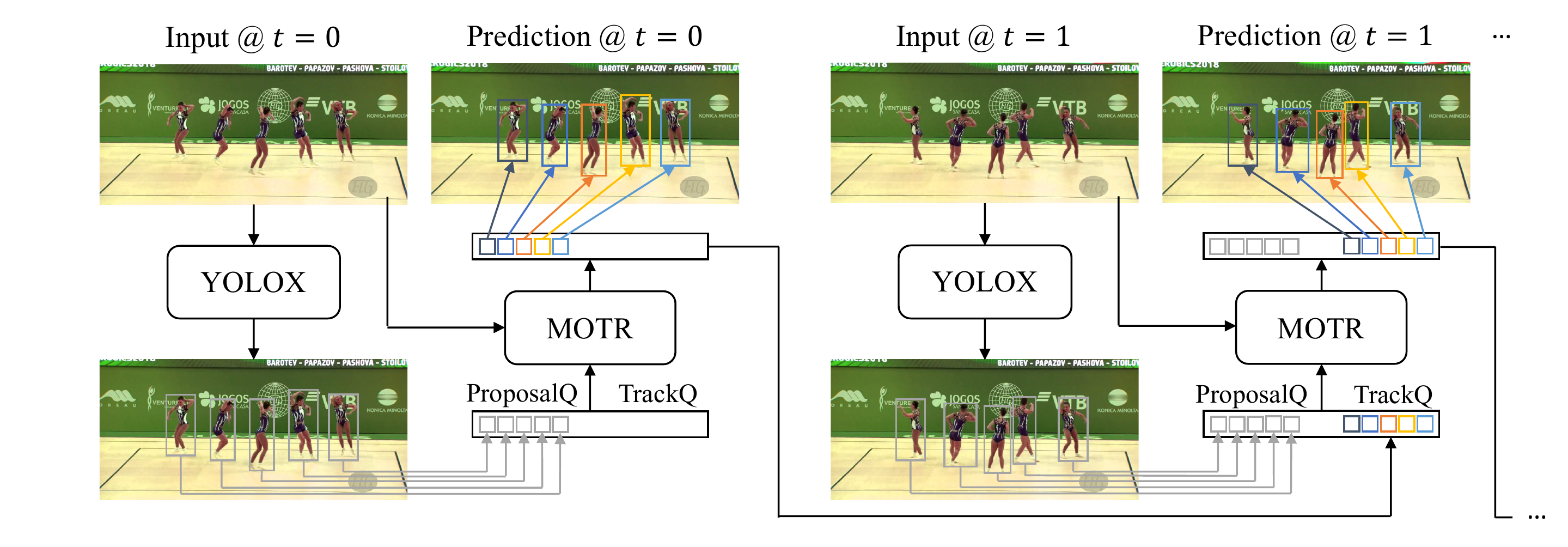}
  \caption{The overall architecture of MOTRv2. The proposals produced by state-of-the-art detector YOLOX \cite{yolox2021} are used to generate the proposal queries, which replaces the detect queries in MOTR~\cite{zeng2022motr} for detecting new-born objects. The track queries are transferred from previous frame and used to predict the bounding boxes for tracked objects. The concatenation of proposal queries and track queries as well as the image features are input to MOTR to generate the predictions frame-by-frame.}
  \label{fig:arch_detail}
\end{figure*}

Inspired by tracking-by-detection methods that take the detection result as the input, we wonder if it is possible to feed the detection result as the input and reduce the learning of MOTR to the association.
%
Recently, there are some advances~\cite{wang2021anchor, liu2022dab} for anchor-based modeling in DETR. For example, DAB-DETR initializes object queries with the center points, height, and width of anchor boxes. Similar to them, we modify the initialization of both detect and track queries in MOTR. We replace the learnable positional embedding (PE) of detect query in MOTR with the sine-cosine PE~\cite{vaswani2017attention} of anchors, producing an anchor-based MOTR tracker. With such anchor-based modeling, proposals generated by an extra object detector can serve as the anchor initialization of MOTR, providing local priors. The transformer decoder is used to predict the relative offsets \wrt the anchors, making the optimization of the detection task much easier.

The proposed MOTRv2 brings many advantages compared to the original MOTR. It greatly benefits from the good detection performance introduced by the extra object detector. The detection task is implicitly decoupled from the MOTR framework, easing the conflict between the detection and association tasks in the shared transformer decoder. MOTRv2 learns to track the instances across frames given the detection results from an extra detector.

MOTRv2 achieves large performance improvements on the DanceTrack, BDD100K, and MOT17 datasets compared to the original MOTR (see Fig.~\ref{fig:comparewithv1}).
On the DanceTrack dataset, MOTRv2 surpasses the tracking-by-detection counterparts by a large margin (\textbf{14.8\%} HOTA compared to OC-SORT~\cite{cao2022observation}), and the AssA metric is \textbf{18.8\%} higher than the second-best method.
On the large-scale multi-class BDD100K dataset~\cite{bdd100k}, we achieved 43.6\% mMOTA, which is 2.4\% better than the previous best solution Unicorn \cite{unicorn}.
MOTRv2 also achieves state-of-the-art performance on the MOT17 dataset~\cite{leal2015motchallenge,milan2016mot16}. We hope our simple and elegant design can serve as a strong baseline for future end-to-end multi-object tracking research.

\section{Related Works}
\label{sec:related}

\para{Tracking by Detection}
Predominant approaches~\cite{zhang2021bytetrack,cao2022observation} mainly follow the tracking-by-detection pipeline:
an object detector first predicts the object bounding boxes for each frame, and a separate algorithm is then used to associate the instance bounding boxes across adjacent frames. The performance of these methods greatly depends on the quality of object detection.

There are various attempts using the Hungarian algorithm~\cite{kuhn1955hungarian} for association:
SORT~\cite{bewley2016simple} applies a Kalman filter~\cite{welch1995introduction} for each tracked instance and uses the intersection-over-union (IoU) matrix among the predicted boxes of Kalman filter and the detected boxes for matching.
DeepSORT~\cite{wojke2017simple} introduces a separate network to extract the appearance features of the instances and uses the pairwise cosine distances on top of SORT.
JDE~\cite{wang2019jde}, Track-RCNN~\cite{shuai2020multi}, FairMOT~\cite{zhang2020fairmot}, and Unicorn \cite{unicorn} further explores the joint training of object detection and appearance embedding.
ByteTrack~\cite{zhang2021bytetrack} leverages a powerful YOLOX-based~\cite{yolox2021} detector and achieves state-of-the-art performance. It introduces an enhanced SORT algorithm to associate the low score detection boxes as well instead of only associating the high score ones.
BoT-SORT~\cite{aharon2022bot} further designs a better Kalman filter state, camera-motion compensation, and ReID feature fusion.
TransMOT~\cite{chu2021transmot} and GTR~\cite{zhou2022global} employ spatial-temporal transformers for instance feature interaction and historical information aggregation when calculating the assignment matrix.
OC-SORT~\cite{cao2022observation} relaxes the linear motion assumption and uses a learnable motion model.

While our approach also benefits from a robust detector, we do not compute similarity matrices but use track queries with anchors to jointly model motion and appearance.

\para{Tracking by Query Propagation}
Another paradigm of MOT extends query-based object detectors~\cite{carion2020detr, zhu2020deformdetr, sun2020sparsercnn} to tracking.
These methods force each query to recall the same instance across different frames.
The interaction between the query and image feature can be performed in parallel or serially in time.

The \emph{parallel} methods take a short video as input and use a set of queries to interact with all frames to predict the trajectories of instances.
VisTR~\cite{vistr2021} and subsequent works~\cite{wu2021seqformer, cheng2021mask2former} extend DETR~\cite{carion2020detr} to detect tracklets in short video clips.
Parallel methods need to take the entire video as input, so they are memory-consuming and limited to short video clips of a few dozen frames.

The \emph{serial} methods perform frame-by-frame query interaction with image features and iteratively update the track queries associated with the instances.
Tracktor++~\cite{bergmann2019twb} utilizes the R-CNN~\cite{girshick2014rich} regression head for iterative instance re-localization across frames.
TrackFormer~\cite{Meinhardt2021trackformer} and MOTR~\cite{zeng2022motr} extend from the Deformable DETR~\cite{zhu2020deformdetr}. They predict the object bounding boxes and update the tracking query for detecting the same instances in subsequent frames.
%
%
%
MeMOT~\cite{cai2022memot} builds the short-term and long-term instance feature memory banks to generate the track queries.
%
TransTrack \cite{transtrack} propagates track queries once to find the object location in the following frame. P3AFormer \cite{zhao2022tracking} adopts flow-guided image feature propagation. Unlike MOTR, TransTrack and P3AFormer still use location-based Hungarian matching in historical tracks and current detections, rather than propagating queries throughout the video.

Our approach inherits the query propagation method for long-term end-to-end tracking, while also utilizing a powerful object detector to provide object location prior. The proposed method greatly outperforms the existing matching and query-based methods in terms of tracking performance in complex motions.

\section{Method}
\label{sec:method}
Here, we present MOTRv2 based on proposal query generation (Sec.~\ref{sec:proposal_query_generation}) and proposal propagation (Sec.~\ref{sec:anchor_propagation}).

\subsection{Revisiting MOTR}

MOTR~\cite{zeng2022motr} is a fully end-to-end multiple-object tracking framework built upon the Deformable DETR \cite{zhu2020deformdetr} architecture. It introduces the track query and object query. The object query is responsible for detecting new-born or missed objects, while each track query is responsible for tracking a unique instance over time. To initialize track queries, MOTR uses the output of the object query associated with newly detected objects. Track queries are updated by their state and current image features over time, which allows them to predict tracks in an online manner.

The tracklet-aware label assignment in MOTR assigns track queries to their previously tracked instances while assigning object queries to the remaining instances by bipartite matching.
MOTR introduces a temporal aggregation network to enhance the features of track queries, and a collective average loss to balance the loss across frames.

\subsection{Motivation}
One major limitation of the end-to-end multiple-object tracking frameworks is their poor detection performance, compared to tracking-by-detection approaches~\cite{zhang2021bytetrack,cao2022observation} that rely on standalone object detectors.
To address this limitation, we propose to incorporate the YOLOX~\cite{yolox2021} object detector to generate proposals as object anchors, providing detection prior to MOTR. It greatly eases the conflict between joint learning detection and association tasks in MOTR, improving the detection performance.


\subsection{Overall Architecture}
As shown in \autoref{fig:arch_detail}, the proposed MOTRv2 architecture consists of two main components: a state-of-the-art object detector and a modified anchor-based MOTR tracker.

The object detector component first generates proposals for both training and inference. For each frame, YOLOX generates a set of proposals that include center coordinates, width, height, and confidence values. The modified anchor-based MOTR component is responsible for learning track association based on the generated proposals.
Sec.~\ref{sec:proposal_query_generation} describes the replacement of the detect queries in the original MOTR framework with proposal queries. The modified MOTR now takes the concatenation of the track query and proposal query as input. Sec.~\ref{sec:anchor_propagation} describes the interaction between concatenated queries and frame features to update the bounding boxes of tracked objects.





\subsection{Proposal Query Generation}
\label{sec:proposal_query_generation}

In this section, we explain how the proposal query generation module provides MOTR with high-quality proposals from YOLOX. The input to this module is a set of proposal boxes generated by YOLOX for each frame in the video. Unlike DETR~\cite{carion2020detr} and MOTR, which use a fixed number of learnable queries for object detection, our framework dynamically determines the number of proposal queries based on the selected proposals generated by YOLOX.

\begin{figure}
  \centering
  \includegraphics[width=\linewidth]{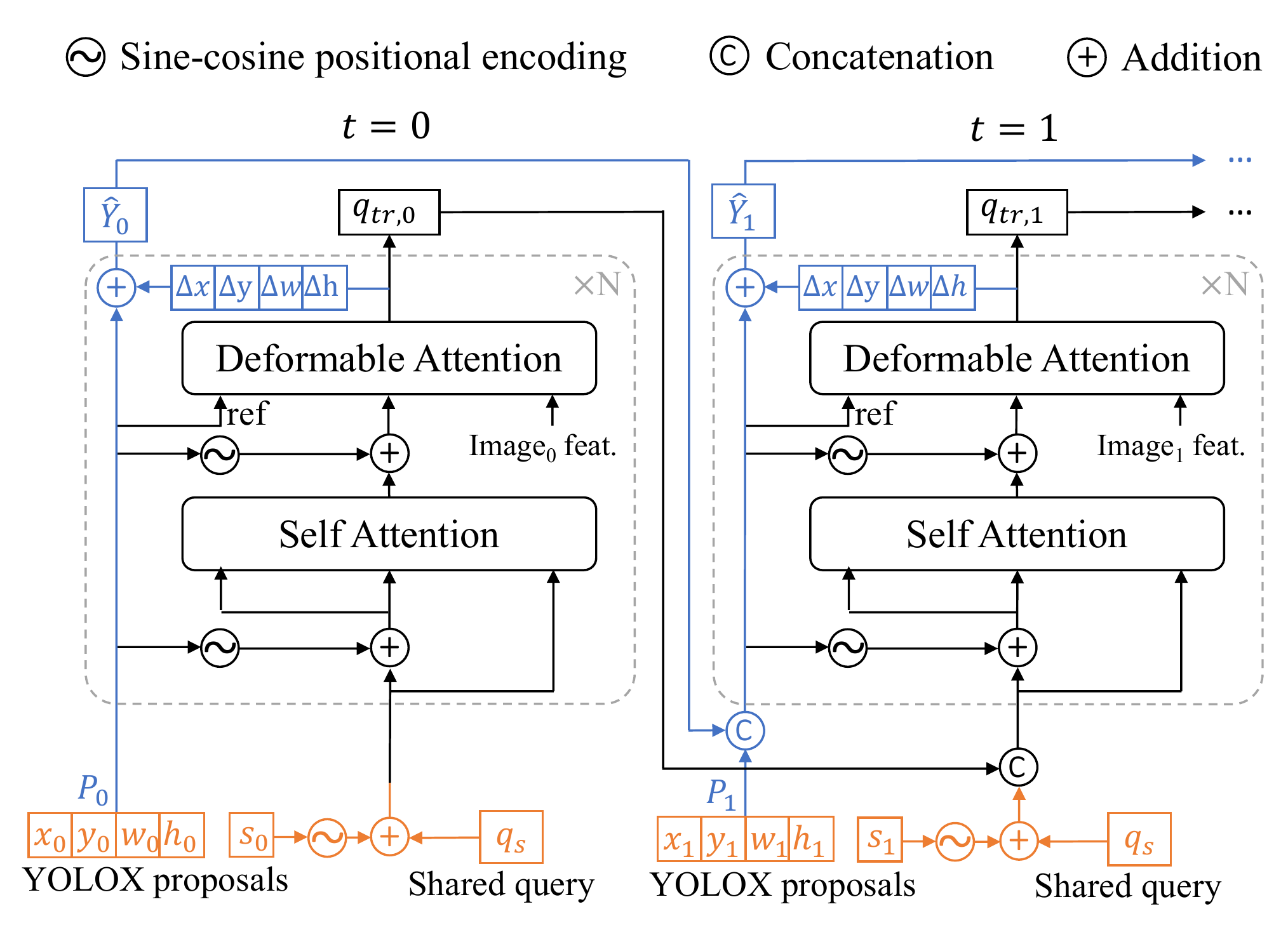}
  \caption{Proposal query generation and proposal propagation for Tracking. The orange color marks proposal query generation while the blue color marks the proposal propagation path; the dashed gray box stands for $N$ transformer decoders. The query interaction module in MOTR is omitted for simplicity.}
  \label{fig:anchor_prop}
\end{figure}

Specifically, for frame $t$, YOLOX generates $N_t$ proposals, each represented by a bounding box with center coordinates ($x_t$, $y_t$), height $h_t$, width $w_t$, and confidence score $s_t$.
As illustrated in the orange part of \autoref{fig:anchor_prop}, we introduce a shared query $q_{s}$ to generate a set of proposal queries.
The shared query, which is a learnable embedding of size $1 \times D$, is first broadcasted to the size of $N_t\times D$. The predicted scores $s_{t}$ of $N_t$ proposal boxes produce the score embeddings of size $N_t\times D$ by sine-cosine positional encoding. The broadcasted queries are then added with the score embeddings to generate the proposal queries.
The YOLOX proposal boxes serve as the anchors of the proposal queries.
In practice, we also use 10 learnable anchors (similar to DAB-DETR~\cite{liu2022dab}) and concatenate them with YOLOX proposals to recall objects missed by the YOLOX detector.

\subsection{Proposal Propagation}
\label{sec:anchor_propagation}

In MOTR \cite{zeng2022motr}, track query and detect query are concatenated and input to the transformer decoder for simultaneous object detection and track association. The track queries generated from the previous frame represents the tracked objects, which are used to predict the bounding boxes of current frame. The detect queries is a fixed set of learnable embeddings and used to detect the new-born objects. Different from MOTR, our method uses the proposal query for detecting new-born objects and the prediction of track queries are made based on previous frame prediction.

For the first frame ($t=0$), there are only new-born objects, which are detected by the YOLOX. As mentioned above, the proposal queries are generated given the shared query $q_{s}$ and predicted scores of YOLOX proposals. After the positional encoding by YOLOX proposals $P_{0}$, the proposal queries are further updated by the self-attention and interact with the image features by deformable attention to produce the track queries $q_{tr,0}$ and the relative offsets $(\Delta x,\Delta y, \Delta w, \Delta h)$ \wrt to the YOLOX proposals $P_{0}$. The prediction $\hat{Y}_{0}$ is the sum of the proposals $P_{0}$ and the predicted offsets.

For other frames ($t>0$), similar to MOTR, track queries $q_{tr,t-1}$ generated from the previous frame will be concatenated with the proposal queries $q_{p,t}$ of the current frame.
The box predictions $\hat{Y}_{t-1}$ of the previous frame will also be concatenated together with the YOLOX proposals $P_{t}$ to serve as the anchors for the current frame. The sine-cosine encoding of the anchors is used as the positional embedding for the concatenated queries, which then go to the transformer decoder to produce the prediction and updated track queries.
The bounding box prediction consists of confidence scores and relative offsets \wrt anchors, and the updated track queries $q_{tr,t}$ are further transferred to the next frame for detecting tracked objects.
%

\begin{figure}
  \includegraphics[width=\linewidth]{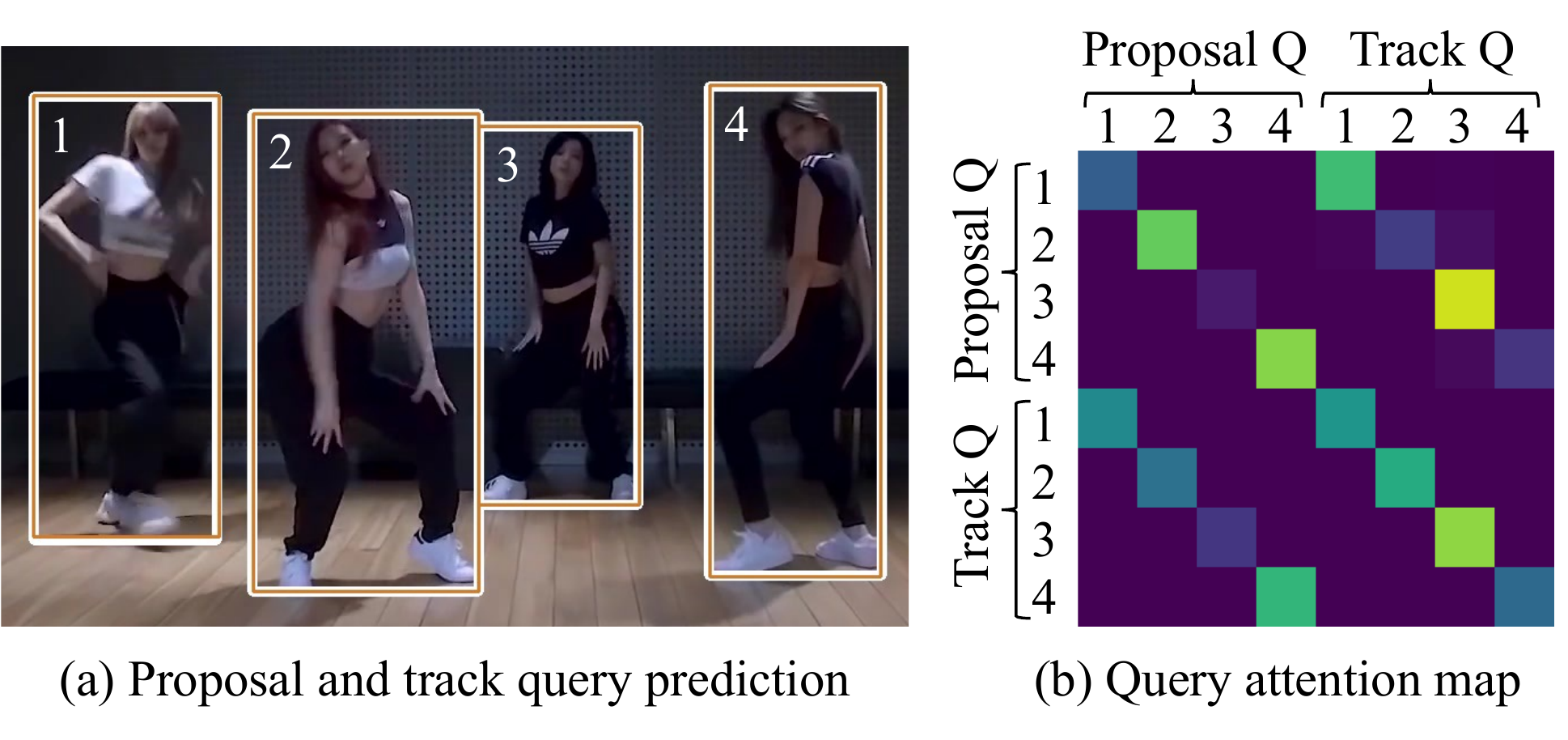}
  \caption{Visualization of (a) MOTR track query box prediction (brown boxes) highly overlaps the YOLOX proposals (while boxes in bold) on the 100th frame of sequence ``dancetrack0005'', and (b) the query self-attention map shows a clear exchange of information between the proposal query and the corresponding tracking query of the same instance.}
  \label{fig:query_attn}
\end{figure}

\para{Analysis}
In the above design, the proposals queries are constrained to detect only the new-born or missing objects and the track queries are responsible for relocating the tracked objects. The proposal queries need to aggregate information from track queries to avoid duplicated detection of tracked objects, and track queries can utilize YOLOX proposals to improve object localization. This is accomplished by the self-attention layers in the transformer decoder.
To better understand the interaction between proposal queries and track queries, we visualize the query self-attention map in \autoref{fig:query_attn}. For the same instance, the proposal query and the corresponding track query have high similarity, and there is a clear exchange of information between them, which verifies our assumptions.


\section{Experiments}

\subsection{Datasets and Metrics}

\para{Datasets} We use the DanceTrack \cite{peize2021dance}, MOT17 \cite{leal2015motchallenge,milan2016mot16} and BDD100K \cite{bdd100k} datasets to evaluate our approach.

\emph{DanceTrack} \cite{peize2021dance} is a large-scale dataset for multi-human tracking in dancing scenes. It features a uniform appearance and diverse motion which is challenging for associating instances across frames. DanceTrack includes 100 videos: 40 for training, 25 for validation, and 35 for testing. The average length of videos is 52.9s.

\emph{MOT17} \cite{leal2015motchallenge,milan2016mot16} is a widely used dataset containing 7 sequences for training and 7 for testing. It mainly contains relatively crowded street scenes with simple and linear movement of pedestrians.

\emph{BDD100K} \cite{bdd100k} is a dataset of autonomous driving scenarios. It contains a multi-object tracking subset with 1400 sequences for training and 200 sequences for validation. The sequence length is about 40 seconds and the number of object classes is 8. We use it to test the multi-class multi-object tracking performance.

\para{Metrics} We use the higher order metric for multi-object tracking (Higher Order Tracking Accuracy; HOTA) \cite{hota2021} to evaluate our method and analysis the contribution decomposed into detection accuracy (DetA) and association accuracy (AssA). For MOT17 and BDD100K datasets, we list the MOTA \cite{bernardin2008evaluating} and IDF1 \cite{ristani2016performance} metrics.

\subsection{Implementation Details}
\label{sec:impl}

\para{Proposal Generation}
We use the YOLOX \cite{yolox2021} detector with weights provided by ByteTrack~\cite{zhang2021bytetrack} and DanceTrack~\cite{peize2021dance} to generate object proposals.
The hyperparameters, such as the input image size, are consistent with ByteTrack. To maximize the proposal recall, we keep all YOLOX predicted boxes with confidence scores over 0.05 as proposals. For DanceTrack \cite{peize2021dance}, we use the YOLOX weights from the DanceTrack official GitHub repository\footnote{https://github.com/DanceTrack/DanceTrack}.
For CrowdHuman~\cite{shao2018crowdhuman} and MOT17, we use the public weight for the MOT17 test set from ByteTrack~\cite{zhang2021bytetrack}. We do not train YOLOX on these two datasets and only use it to generate proposals for all images before training MOTR.
For BDD100K \cite{bdd100k}, we use its MOT set together with the 100k images set for training. The YOLOX detector is trained on 8 Tesla V100 GPUs for 16 epochs. We follow ByteTrack \cite{zhang2021bytetrack} for other training hyperparameters.

\para{MOTR}
Our implementation is based on the official repo\footnote{https://github.com/megvii-research/MOTR} with a ResNet50~\cite{He2016Resnet} backbone for feature extraction. All MOTR models are trained on 8 GPUs, with a batch size of 1 per GPU.
For DanceTrack \cite{peize2021dance}, we follow YOLOX \cite{yolox2021} and adopt the HSV augmentation for training MOTR.
As opposed to the original implementation of propagating track queries that match ground truth tracks during training, we propagate track queries with a confidence score above 0.5, which naturally creates false positive (FP; high score but no instances, \eg lost tracks) and false negative (FN; instances not detected) track queries to enhance the handling of FPs and FNs during inference. In this way, we do not follow MOTR to manually insert negative or drop positive track queries, \ie, $p_{drop}=0$ and $p_{insert}=0$.
We train the ablation study and state-of-the-art comparison models for 5 epochs with a fixed clip size of 5. The sampling stride of the frames inside a clip is randomly chosen from 1 to 10. The initial learning rate $2\times 10^{-4}$ is dropped by a factor of 10 at the 4$\rm ^{th}$ epoch.
For MOT17 \cite{leal2015motchallenge,milan2016mot16}, the number of training epochs is tuned down to 50 and the learning rate drops at the 40$\rm ^{th}$ epoch.
For BDD100K \cite{bdd100k}, we train for 2.5 epochs using a clip size of 4 with a random sampling stride from 1 to 4. The learning rate drops at the end of the 2$\rm ^{nd}$ epoch.
For the extension to multi-class MOT, each YOLOX proposal additionally includes a class label and we use a different learnable embedding (shared query) for each class.
Other settings are not changed.


\para{Joint Training with CrowdHuman}
To improve the detection performance, we also utilize a large number of static CrowdHuman images. For the DanceTrack dataset, similar to the joint training of MOT17 and CrowdHuman in MOTR, we generate pseudo-video clips for CrowdHuman and perform joint training with DanceTrack. The length of the pseudo-video clips is also set to 5.
We use the 41,796 samples from the training set of the DanceTrack \cite{peize2021dance} dataset and the 19,370 samples from the training and validation set of the CrowdHuman \cite{shao2018crowdhuman} dataset for joint training.
%
For the MOT17 Dataset, we keep the original settings in MOTR that concatenate the validation set of CrowdHuman and the training set of MOT17.

\subsection{State-of-the-art Comparison on DanceTrack}

\begin{table}[t]
  \centering
  \caption{Performance comparison with state-of-the-art methods on the DanceTrack\cite{peize2021dance} test set. Results for existing methods are from DanceTrack \cite{peize2021dance}. MOTRv2$^*$ denotes MOTRv2 with an extra association, adding validation set for training, and test ensemble.}
  \vspace{-1.6ex}
  \setlength{\tabcolsep}{1.2mm}{
    \begin{tabular}{l|ccccc}
      \toprule
      Methods                                & HOTA     & DetA     & AssA     & MOTA     & IDF1     \\\midrule
      FairMOT \cite{zhang2020fairmot}        & 39.7     & 66.7     & 23.8     & 82.2     & 40.8     \\
      CenterTrack \cite{zhou2020centertrack} & 41.8     & 78.1     & 22.6     & 86.8     & 35.7     \\
      TransTrack \cite{transtrack}           & 45.5     & 75.9     & 27.5     & 88.4     & 45.2     \\
      TraDes \cite{trades2021}               & 43.3     & 74.5     & 25.4     & 86.2     & 41.2     \\
      ByteTrack \cite{zhang2021bytetrack}    & 47.7     & 71.0     & 32.1     & 89.6     & 53.9     \\
      GTR \cite{trades2021}                  & 48.0     & 72.5     & 31.9     & 84.7     & 50.3     \\
      QDTrack \cite{quasisense2021}          & 54.2     & 80.1     & 36.8     & 87.7     & 50.4     \\
      MOTR \cite{zeng2022motr}               & 54.2     & 73.5     & 40.2     & 79.7     & 51.5     \\
      OC-SORT \cite{cao2022observation}      & 55.1     & 80.3     & 38.3     & 92.0     & 54.6     \\ \midrule
      MOTRv2 (ours)                          & 69.9     & 83.0     & 59.0     & 91.9     & 71.7     \\
      MOTRv2$^*$ (ours)                      & \bf 73.4 & \bf 83.7 & \bf 64.4 & \bf 92.1 & \bf 76.0 \\\bottomrule
    \end{tabular}
  }
  \label{tab:dancetrack_sota}
\end{table}

We compare MOTRv2 with the state-of-the-art methods on the DanceTrack \cite{peize2021dance} test set and the results are shown in \autoref{tab:dancetrack_sota}.
Without bells and whistles, our method achieves 69.9 HOTA and shows the best performance on all higher-order metrics, surpassing other state-of-the-art methods by a large margin.
Compared to those matching-based methods, \eg ByteTrack \cite{zhang2021bytetrack} and OC-SORT \cite{cao2022observation}, our approach shows a much better association accuracy (59.0\% \vs 38.3\%) while also achieves decent detection accuracy (83.0\% \vs 80.3\%). MOTRv2 achieves 69.9\% higher order tracking accuracy (HOTA), which is \textbf{14.8\%} better than the previous best method. The large gap in the IDF1 metric between previous methods and MOTRv2 also shows the superiority of our method in complex motions.
For better performance, we apply an extra association in post-processing: if only one track exits and another one appears within 20 to 100 frames, we consider them to be tracks of the same instance. With the extra association, adding the validation set for training, and using an ensemble of 4 models, we further achieve 73.4\% HOTA on the DanceTrack test set.



\subsection{State-of-the-art Comparison on BDD100K}

\begin{table}[t]
  \centering
  \caption{Performance comparison with state-of-the-art methods on the BDD100K\cite{bdd100k} MOT validation set. MOTR$\rm ^*$ means MOTRv2 without using YOLOX proposals.}
  \vspace{-1.6ex}
  \setlength{\tabcolsep}{1.8mm}{
  \begin{tabular}{l|cccc}
    \toprule
    Methods                       & mMOTA & mIDF1 & MOTA & IDF1 \\\midrule
    Yu \etal \cite{bdd100k}       & 25.9  & 44.5  & 56.9 & 66.8 \\
    QDTrack \cite{quasisense2021} & 36.6  & 50.8  & 63.5 & 71.5 \\
    TETer \cite{li2022tracking}   & 39.1  & 53.3  & /    & /    \\
    Unicorn \cite{unicorn}        & 41.2  & 54.0  & \bf 66.6 & 71.3 \\\midrule
    MOTR    \cite{zeng2022motr}   & 32.3  & 44.8  & 56.2 & 65.8 \\
    MOTR$\rm ^*$                  & 35.5  & 48.2  & 59.6 & 68.9 \\
    MOTRv2 (ours)                 & \bf 43.6  & \bf 56.5  & 65.6 & \bf 72.7 \\
    \bottomrule
  \end{tabular}
  }
  \label{tab:bdd100k_sota}
\end{table}

\autoref{tab:bdd100k_sota} shows the results on the BDD100k \cite{bdd100k} tracking validation set. MOTRv2 achieved the highest mMOTA and mIDF1 among all methods.
For a fair comparison, we equip the MOTR with 100k image set joint training and box propagation, denoted as MOTR$\rm ^*$. By utilizing YOLOX proposals, MOTRv2 outperforms MOTR$\rm ^*$ by 8.1\% mMOTA and 8.3\% mIDF1, showing that the YOLOX proposals greatly improve the multi-class detection and tracking performance.
Compared to other state-of-the-art methods, MOTRv2 outperforms the best tracker Unicorn by 2.4\% mMOTA and 1.1\% mIDF1. The higher mMOTA and mIDF1 (averaged among all classes) indicate that MOTRv2 handles the multi-class scenarios better. The difference in overall MOTA (-1.0\%) and IDF1 (+1.4\%) shows that our method is better in terms of association.

\subsection{Comparison on the MOTChallenge}

\begin{table}[t]
  \centering
  \caption{Comparison to existing methods on the MOT17 dataset.}
  \vspace{-1.6ex}
  \setlength{\tabcolsep}{1mm}{
    \begin{tabular}{l|cccccc}
      \toprule
      Methods                                     & HOTA & AssA & DetA & IDF1 & MOTA \\
      \midrule
      \textit{CNN-based:}                         &      &      &      &      &      \\
      Tracktor++\cite{bergmann2019twb}            & 44.8 & 45.1 & 44.9 & 52.3 & 53.5 \\
      CenterTrack\cite{zhou2020centertrack}       & 52.2 & 51.0 & 53.8 & 64.7 & 67.8 \\
      TraDeS \cite{trades2021}                    & 52.7 & 50.8 & 55.2 & 63.9 & 69.1 \\
      QuasiDense \cite{quasisense2021}            & 53.9 & 52.7 & 55.6 & 66.3 & 68.7 \\
      GSDT \cite{wang2021joint}                   & 55.5 & 54.8 & 56.4 & 68.7 & 66.2 \\
      FairMOT\cite{zhang2020fairmot}              & 59.3 & 58.0 & 60.9 & 72.3 & 73.7 \\
      CorrTracker \cite{wang2021multiple}         & 60.7 & 58.9 & 62.9 & 73.6 & 76.5 \\
      Unicorn \cite{unicorn}                      & 61.7 & /    & /    & 75.5 & 77.2 \\
      GRTU \cite{wang2021general}                 & 62.0 & 62.1 & 62.1 & 75.0 & 74.9 \\
      MAATrack \cite{stadler2022modelling}        & 62.0 & 60.2 & 64.2 & 75.9 & 79.4 \\
      ByteTrack \cite{zhang2021bytetrack}         & 63.1 & 62.0 & 64.5 & 77.3 & 80.3 \\
      OC-SORT \cite{cao2022observation}           & 63.2 & 63.2 & /    & 77.5 & 78.0 \\
      BoT-SORT \cite{aharon2022bot}               & 64.6 & /    & /    & 79.5 & 80.6 \\
      \midrule
      \textit{Transformer-based:}                 &      &      &      &      &      \\
      TrackFormer \cite{Meinhardt2021trackformer} & /    & /    & /    & 63.9 & 65.0 \\
      TransTrack\cite{transtrack}                 & 54.1 & 47.9 & 61.6 & 63.9 & 74.5 \\
      MOTR \cite{zeng2022motr}                    & 57.8 & 55.7 & 60.3 & 68.6 & 73.4 \\
      P3AFormer \cite{zhao2022tracking}           & /    & /    & /    & 78.1 & 81.2 \\
      MOTRv2 (ours)                               & 62.0 & 60.6 & 63.8 & 75.0 & 78.6 \\
      \bottomrule
    \end{tabular}}
  \label{tab_compare_sota}
\end{table}

\begin{table}[t]
    \centering
    \caption{Comparison to existing methods on the MOT20 test set.}
    \vspace{-1.6ex}
  \setlength{\tabcolsep}{1mm}{
    \begin{tabular}{l|cccccc}
      \toprule
      Methods                                     & HOTA & AssA & DetA & IDF1 & MOTA \\
      \midrule
      FairMOT \cite{zhang2020fairmot}             & 54.6 & 54.7 &	54.7 & 67.3 & 61.8 \\
      ByteTrack \cite{zhang2021bytetrack}         & 61.3 & 59.6 & 63.4 & 75.2 & 77.8 \\
      OC-SORT \cite{cao2022observation}           & 62.4 & 62.5 & / & 76.4 & 75.9 \\
       \hline
      MOTRv2 (ours)                               & 60.3 & 58.1 & 62.9 & 72.2 & 76.2 \\
       + MOT17 joint train & 61.0 & 59.3  & 63.0 & 73.1 & 76.2 \\
      \bottomrule
    \end{tabular}}
    \label{tab:mot20}
\end{table}

We further compare the performance of MOTRv2 with the state-of-the-art methods on the MOT17 \cite{leal2015motchallenge,milan2016mot16} and MOT20 \cite{dendorfer2020mot20} datasets.
\autoref{tab_compare_sota} shows the comparison on MOT17. Compared to the original MOTR \cite{zeng2022motr}, the introduction of YOLOX proposals consistently improves the detection (DetA) and association (AssA) accuracy by 3.5\% and 4.9\% correspondingly.
The proposed approach pushes the performance of query-based trackers in crowded scenarios to the state-of-the-art level. We attribute the remaining performance gap to the fact that the scale of the MOT17 dataset is too small (215 seconds in total), which is insufficient to train a query-based tracker.
\autoref{tab:mot20} shows our result on the MOT20 \cite{dendorfer2020mot20} dataset. The performance gap between our method and ByteTrack~\cite{zhang2021bytetrack} can be reduced with the joint training of MOT17, especially for the AssA metric. It also suggests that the lower performance in the MOT challenge is more likely due to the small size of real videos.

\subsection{Ablation Study}
In this section, we study several components of our method, including YOLOX proposals, proposal propagation, and CrowdHuman joint training. \autoref{tab:hota_summary} summarizes the effect of components on DanceTrack validation and test sets. The improvements are consistent across both sets.

\begin{table}[t]
    \small
    \centering
    \caption{Summary of cumulative improvements on DanceTrack.}
    \vspace{-2ex}
    \setlength{\tabcolsep}{1.8mm}{
    \begin{tabular}{lcc}
    \toprule
         & val HOTA & test HOTA \\
    \midrule
        MOTR [\href{https://arxiv.org/abs/2105.03247}{1}]       & 51.7 & 54.2 \\
        + Implementation (Sec.~\ref{sec:impl} MOTR)   & 54.8 & / \\
        + Propagate boxes & 57.1 & / \\
        + YOLOX \& CrowdHuman (Tab. \ref{tab:yolox}) & 63.7 & / \\
        + Query denoise (Tab. \ref{tab:qd})       & 64.5 & 69.9 \\
        \multicolumn{2}{l}{+ Extra association, val set, test ensemble} & 73.4 \\
    \bottomrule
    \end{tabular}}
    \label{tab:hota_summary}\vspace{-0.8em}
\end{table} 

\para{YOLOX Proposal}
For a more thorough study of the benefits of using the YOLOX proposal, we test the effect of YOLOX proposals under two settings: with and without CrowdHuman joint training.
\autoref{tab:yolox} shows that using YOLOX predictions as proposal queries \emph{consistently improves all three metrics} (HOTA, DetA, and AssA) regardless of whether the CrowdHuman dataset is used. The YOLOX proposals significantly improve association accuracy (AssA) by $9.3\%$ when trained jointly with the CrowdHuman dataset.
Using the pretrained object detector YOLOX alone outperforms that of joint training with the CrowdHuman dataset (HOTA 56.7 \vs 60.7).

\begin{table}
  \centering
  \caption{Ablation study of CrowdHuman joint training and YOLOX proposal on the DanceTrack validation set.}
  \vspace{-2ex}
  \label{tab:yolox}
  \begin{tabular}{cc|ccc}
    \toprule
    CrowdHuman   & YOLOX        & HOTA     & DetA     & AssA     \\ \midrule
                 &              & 57.1     & 66.2     & 49.5     \\
                 & $\checkmark$ & 60.7     & 74.8     & 49.6     \\
    $\checkmark$ &              & 56.7     & 73.7     & 43.9     \\
    $\checkmark$ & $\checkmark$ & \bf 63.7 & \bf 76.6 & \bf 53.2 \\ \bottomrule
  \end{tabular}
\end{table}

%

Both using YOLOX proposals and CrowdHuman joint training improve detection accuracy as expected. However, using CrowdHuman pseudo-videos seems to have a negative impact on the training of association, as indicated by the 5.6\% drop in AssA. This might be caused by the gap between the two datasets: the CrowdHuman pseudo-videos bias the training towards enabling learnable detect queries to handle more difficult detections, and the human motion of pseudo-videos created by affine transformations are different from that of DanceTrack.
It is worth noticing that using YOLOX proposals in turn helps CrowdHuman joint training.
Our method of using YOLOX proposals makes detection easier for MOTR, thus alleviating the bias toward detection and the conflict between the detection and association tasks. As a result, with the YOLOX proposals, joint training with CrowdHuman can further improve rather than hurt the tracking performance.

\para{Proposal Propagation}
Here, we show the effect of propagating proposals (center point as well as width and height) from the current frame to the subsequent frame. The baseline for comparison is the propagation of reference point as applied in MOTR~\cite{zeng2022motr} and TransTrack \cite{transtrack}. It means that only the center point from the previous frame is employed as the query reference point. In addition, we explore the effect of replacing the learnable positional embedding of queries with the sine-cosine positional encoding of anchors (or reference points).
\begin{table}
  \centering
  \caption{Ablation study on propagating anchors \vs center points and learnable \vs sine-cosine positional encoding.}
  \vspace{-2ex}
  \label{tab:anchor_prop}
  \begin{tabular}{ll|ccc}
    \toprule
    Propagate & Box embedding & HOTA & DetA & AssA \\ \midrule
    Point     & Learnable     & 61.2 & 76.6 & 49.2 \\
    Point     & Sine-cosine   & 60.5 & 76.6 & 48.0 \\
    Box       & Learnable     & 63.8 & 76.9 & 53.1 \\
    Box       & Sine-cosine   & 63.7 & 76.6 & 53.2 \\ \bottomrule
  \end{tabular}
\end{table}
From \autoref{tab:anchor_prop}, we can easily find that propagating four-dimensional proposals (boxes) instead of center points yields much better association performance. It indicates that MOTRv2 benefits from the width and height information from the bounding box prediction of the previous frame for associating instances.
In contrast, the sine-cosine positional encoding barely helps association compared to the original design of using learnable positional embedding in Deformable DETR \cite{zhu2020deformdetr}. Therefore, using the anchor boxes instead of points is not only critical for introducing YOLOX detection results but also sufficient for providing the MOTR decoder with localization information.

\para{Score Encoding}
As mentioned in sec.~\ref{sec:proposal_query_generation}, the proposal queries are the sum of two parts: (1) encoding of the confidence scores; (2) a shared learnable query embedding.
We explored two ways to encode the confidence score of YOLOX proposals, namely linear projection and sine-cosine positional encoding.
For linear projection, we use a simple weight matrix of size $1 \times D$ to expand the score scalar to a $D$-dimensional score embedding.
Additionally, we test not using confidence scores at all, \ie, only using a shared query embedding for proposal queries.
\begin{table}
  \centering
  \caption{Effect of using the confidence score of YOLOX proposals and different methods for encoding confidence score.}
  \vspace{-2ex}
  \label{tab:yolox_score}
  \begin{tabular}{l|ccc}
    \toprule
    Score embedding   & HOTA     & DetA & AssA \\ \midrule
    Not applied       & 63.0     & 77.3 & 51.5 \\
    Linear projection & 63.4     & 77.4 & 52.1 \\
    Sine-cosine       & \bf 63.7 & 76.6 & 53.2 \\ \bottomrule
  \end{tabular}
\end{table}
\autoref{tab:yolox_score} shows that not using score embedding performs the worst, which means the confidence score provides important information for MOTR.
Further, both learnable embedding and sine-cosine encoding work well, and using sine-cosine encoding works better for the association.

\para{Query Denoising}
For fast convergence in training, we introduce query denoising~\cite{li2022dn} (QD) as an auxiliary task for DanceTrack and MOT17.
\autoref{tab:qd} shows that query denoising with the default noise scale (0.4) hurts the association performance. We attribute this to the gap between detection and tracking, as artificial noise is usually larger in scale compared to the cross-frame motion of instances. Our choice of noise range achieves a 2.1\% improvement in DetA. Query denoising improves the detection performance and further improves the HOTA metric by $0.8\%$.

\begin{table}
  \centering
  \caption{Effect of query denoising on the DanceTrack validation set. The definition of noise scale $\lambda_1, \lambda_2$ follows DN-DETR \cite{li2022dn}.}
  \vspace{-2ex}
  \label{tab:qd}
  \begin{tabular}{cc|ccc}
    \toprule
    $\lambda_1$               & $\lambda_2$ & HOTA     & DetA     & AssA \\ \midrule
    \multicolumn{2}{c|}{No QD} & 63.7        & 76.6     & \bf 53.2        \\
    0.4                       & 0.4         & 63.1     & 77.7     & 51.5 \\
    0.1                       & 0.05        & \bf 64.5 & \bf 78.7 & 53.0 \\ \bottomrule
  \end{tabular}
\end{table}

\begin{figure}
  \centering
  \includegraphics[width=\linewidth]{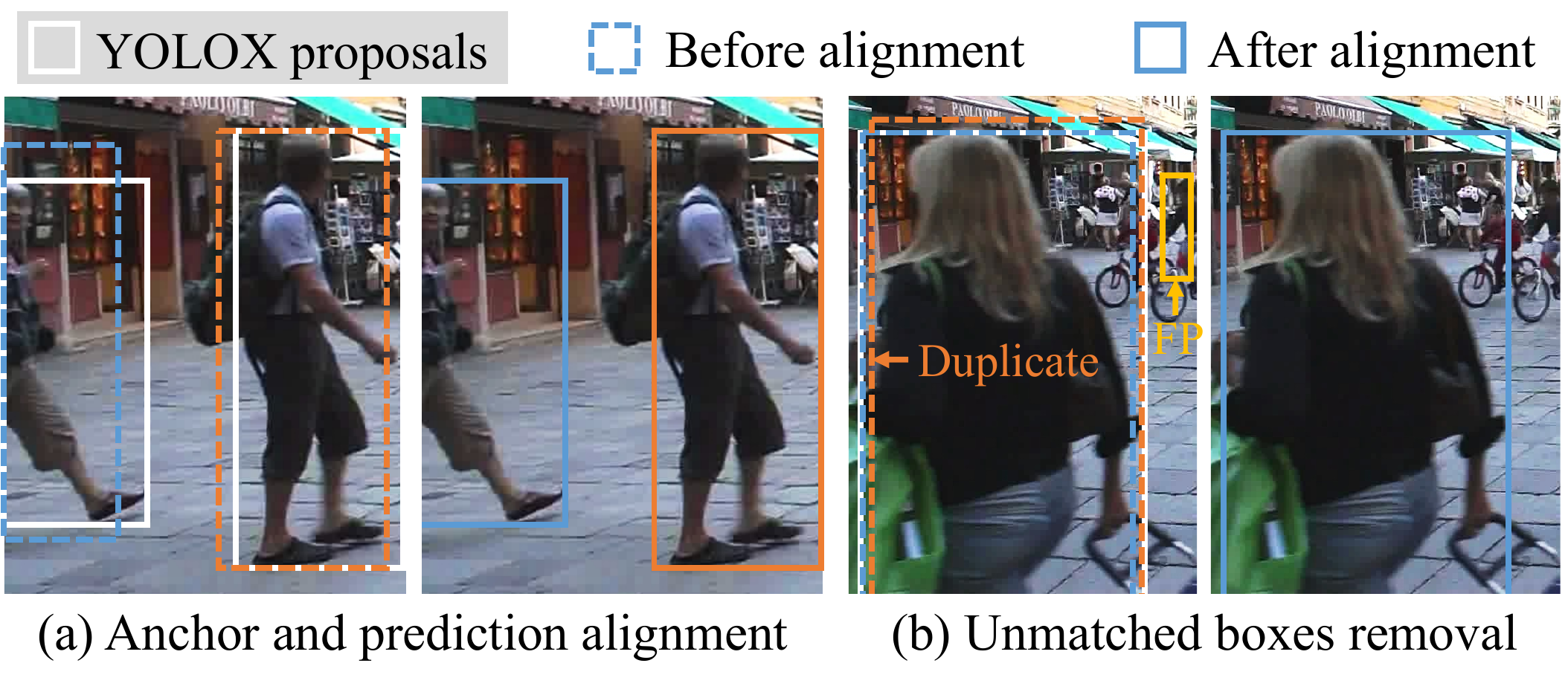}
  \vspace{-4ex}
  \caption{Illustration of Track Query Alignment: (a) imprecise MOTR localization is replaced with corresponding YOLOX proposal boxes for better prediction and anchor localization; (b) false positive detections and duplicate track queries can be eliminated by removing the unmatched boxes.}
  \vspace{-2ex}
  \label{fig:align_result}
\end{figure}

\para{Track Query Alignment}
To take full advantage of accurate object detection from YOLOX in crowd scenarios, we further introduce track query alignment for enhancing MOTRv2 specifically on the MOT17 \cite{leal2015motchallenge,milan2016mot16} and MOT20 \cite{dendorfer2020mot20} datasets.
We first calculate the intersection-over-union (IoU) matrix between the MOTR-predicted boxes and YOLOX proposals. Then, we perform Hungarian matching on the IoU matrix to find the best-matched pairs and keep all matched pairs of boxes with an IoU over 0.5. After that, we propose three independent alignment strategies: the matched YOLOX boxes can replace (1) MOTR box \emph{predictions} of that frame and (2) the track query \emph{anchors} for detecting the corresponding instances in the next frame. Further, (3) unmatched MOTR predictions can be \emph{removed} from prediction as they are likely to be false positives.
\autoref{fig:align_result} illustrate the effects of these alignments. Note that these alignments only apply to anchor or prediction boxes and do not change the propagation of query embedding, which preserves the end-to-end nature of our method.

\begin{table}
  \centering
  \caption{Effect of track query alignment on MOT17 valhalf.}
  \vspace{-2ex}
  \label{tab:alignment}
  \begin{tabular}{ccc|cc}
    \toprule
    Prediction   & Anchor       & Removal      & MOTA     & IDF1     \\ \midrule
                 &              &              & 75.9     & 75.1     \\
                 &              & $\checkmark$ & 83.0     & 76.7     \\
                 & $\checkmark$ &              & 84.3     & 79.0     \\
                 & $\checkmark$ & $\checkmark$ & 86.3     & 79.5     \\
    $\checkmark$ &              &              & 77.0     & 75.4     \\
    $\checkmark$ &              & $\checkmark$ & 84.1     & 77.0     \\
    $\checkmark$ & $\checkmark$ &              & 84.9     & 79.2     \\
    $\checkmark$ & $\checkmark$ & $\checkmark$ & \bf 86.8 & \bf 79.7 \\ \bottomrule
  \end{tabular}
\end{table}


We test the three methods on MOT17 using the first half of each training sequence for training and the remaining for validation. All alignments are applied during training and the ablation study of alignment methods is performed during inference. The results are shown in \autoref{tab:alignment}. Among the three methods, aligning anchors is the most beneficial for detection and tracking performance, as it boosts MOTA by 8.4\% and IDF1 by 3.9\% when used alone (row 1 \vs row 3). Aligning anchors with the corresponding YOLOX proposals mitigates the accumulation of localization errors during anchor propagation, thereby improving both detection and association accuracy (see \autoref{fig:align_result}(a)).
Removing MOTR predictions that do not match any YOLOX boxes can improve detection performance under all settings. It further improves MOTA by 2.0\% in addition to anchor alignment (row 2 \vs row 4)  (see \autoref{fig:align_result}(b)).
Finally, frame-by-frame prediction alignment, as an intuitive approach, can be used to further improve MOTA and IDF1.

\section{Discussion}
In this paper, we propose MOTRv2, an elegant combination of MOTR tracker and YOLOX detector. YOLOX generates high-quality object proposals that help MOTR detect new objects more easily. This reduces the complexity of object detection, allowing MOTR to concentrate on the association process. MOTRv2 breaks through the common belief that end-to-end frameworks are not suitable for high-performance MOT and explains why previous end-to-end MOT frameworks have failed. We hope it can provide some new insights on end-to-end MOT for the community.

\para{Limitations}
Although using the YOLOX proposals greatly ease the optimization problem of MOTR, the proposed method is still data-hungry and does not perform well enough on smaller datasets.
Furthermore, we observe a few duplicated track queries when, for example, when two individuals cross paths with one another. In such cases, one track query might end up following the incorrect subject, leading to two track queries on same individual (see \autoref{fig:align_result}(b)).
This observation could serve as a valuable hint for potential enhancements in the future.
Another limitation is the efficiency. The bottleneck is mainly from the MOTR \cite{zeng2022motr} part. Quantitatively, the YOLOX~\cite{yolox2021} detector runs at 25 FPS while MOTR runs at 9.5 FPS on 2080Ti. Adding these two components yields a speed of 6.9 FPS.

{\small
    \bibliographystyle{ieee_fullname}
    \bibliography{egbib}
  }

\end{document}